\documentclass[sigconf]{acmart}
\usepackage[pagewise]{lineno}
\usepackage{algorithm, algorithmic}
\usepackage{subfigure}
\usepackage{multirow}
\usepackage{graphicx}
\usepackage{utfsym}
\usepackage{url}
\usepackage{balance}
\AtBeginDocument{%
 }

\copyrightyear{2024}
\acmYear{2024}
\setcopyright{rightsretained}
\acmConference[UAVM '24]{Proceedings of the 2nd Workshop on UAVs in Multimedia: Capturing the World from a New Perspective}{October 28-November 1, 2024}{Melbourne, VIC, Australia}
\acmBooktitle{Proceedings of the 2nd Workshop on UAVs in Multimedia: Capturing the World from a New Perspective (UAVM '24), October 28-November 1, 2024, Melbourne, VIC, Australia}
\acmDOI{10.1145/3689095.3689103}
\acmISBN{979-8-4007-1206-7/24/10}

\makeatletter
\gdef\@copyrightpermission{
 \begin{minipage}{0.3\columnwidth}
 \href{https://creativecommons.org/licenses/by/4.0/}{\includegraphics[width=0.90\textwidth]{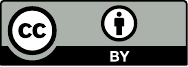}}
 \end{minipage}\hfill
 \begin{minipage}{0.7\columnwidth}
 \href{https://creativecommons.org/licenses/by/4.0/}{This work is licensed under a Creative Commons
Attribution International 4.0 License.}
 \end{minipage}
 \vspace{5pt}
}
\makeatother

\begin{document}
\title{Multi-weather Cross-view Geo-localization Using Denoising Diffusion Models}

\author{Tongtong Feng}
\orcid{0000-0003-4734-5607}
\affiliation{%
  \institution{Department of Computer Science and Technology, Tsinghua University}
  \city{Beijing}
  \country{China}
}
\email{fengtongtong@tsinghua.edu.cn}

\author{Qing Li}
\affiliation{%
  \institution{Department of Electronic Engineering, Tsinghua University}
  \city{Beijing}
  \country{China}
}
\email{soleilor@mail.tsinghua.edu.cn}

\author{Xin Wang}
\authornote{Corresponding Authors. BNRist is the abbreviation of Beijing National Research Center for Information Science and Technology.}
\affiliation{%
  \institution{Department of Computer Science and Technology, BNRist, Tsinghua University, Beijing, China}
  \country{}
}
\email{xin_wang@tsinghua.edu.cn}

\author{Mingzi Wang}
\affiliation{%
  \institution{TBSI, Shenzhen International Graduate School, Tsinghua University}
  \city{Shenzhen}
  \country{China}
}
\email{wmz22@mails.tsinghua.edu.cn}

\author{Guangyao Li}
\affiliation{%
  \institution{Department of Computer Science and Technology, Tsinghua University}
  \city{Beijing}
  \country{China}
}
\email{guangyaoli@tsinghua.edu.cn}

\author{Wenwu Zhu}
\authornotemark[1]
\affiliation{%
  \institution{Department of Computer Science and Technology, BNRist, Tsinghua University, Beijing, China}
  \country{}
}
\email{wwzhu@tsinghua.edu.cn}

\renewcommand{\shortauthors}{Tongtong Feng, Qing Li, Xin Wang, Mingzi Wang, Guangyao Li {\&} Wenwu Zhu.}

\begin{abstract}
Cross-view geo-localization in GNSS-denied environments aims to determine an unknown location by matching drone-view images with the correct geo-tagged satellite-view images from a large gallery. Recent research shows that learning discriminative image representations under specific weather conditions can significantly enhance performance. However, the frequent occurrence of unseen extreme weather conditions hinders progress. This paper introduces MCGF, a Multi-weather Cross-view Geo-localization Framework designed to dynamically adapt to unseen weather conditions. MCGF establishes a joint optimization between image restoration and geo-localization using denoising diffusion models. For image restoration, MCGF incorporates a shared encoder and a lightweight restoration module to help the backbone eliminate weather-specific information. For geo-localization, MCGF uses EVA-02 as a backbone for feature extraction, with cross-entropy loss for training and cosine distance for testing. Extensive experiments on University160k-WX demonstrate that MCGF achieves competitive results for geo-localization in varying weather conditions.
\end{abstract}



\maketitle

\section{Introduction}
\begin{figure}[h]
    \includegraphics[width=0.46\textwidth]{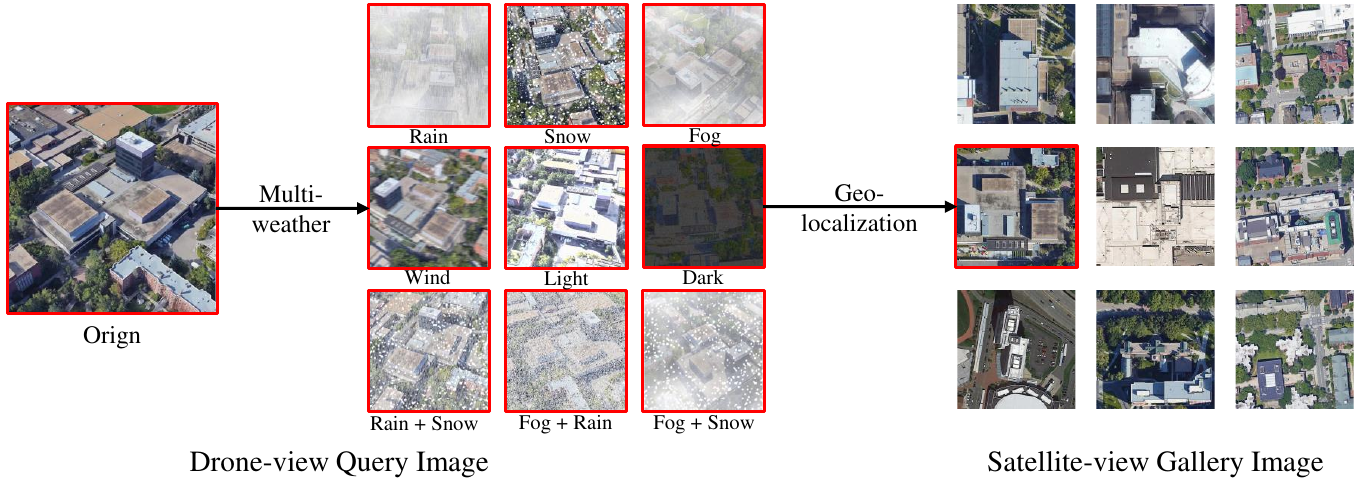}
    \centering
    \caption{Multi-weather cross-view geo-localization. The red box represents the correct match we want to achieve.}
    \label{fig_1}
\end{figure}

Cross-view geo-localization\cite{4} aims to determine an unknown location by matching drone-view images with the correct geo-tagged satellite-view images from a large gallery, based on geographic features in the images, as shown in Figure \ref{fig_1}. This task is crucial for accurate navigation and safe planning\cite{6,7,8} in GNSS-denied autonomous drone flights. Recent advances in vision transformer have led to significant breakthroughs in various cross-view geo-localization tasks, such as drone localization\cite{1,5} (matching drone-view query images with geo-tagged satellite-view images) and drone navigation\cite{2,3} (using satellite-view query images to guide drones to a target area). However, varying weather conditions, including fog, rain, snow, wind, light, dark, and combinations of multiple weather types, reduce visibility, corrupt the information captured by an image, significantly complicate image geographic representation, and lead to a sharp decline in task performance. The major challenge lies in adaptively achieving unbiased image geographic representation under diverse weather conditions.

A clean image without any weather degradation is desired in cross-view geo-localization. Early methods for weather removal using empirical observations \cite{9}, Convolutional Neural Networks (CNNs) based and transformer-based for deraining\cite{10}, dehazing\cite{11}, and desnowing\cite{12}. Most of these methods achieve excellent performance, but these are not generic solutions for all adverse weather removal problems as the networks have to be trained separately for each weather\cite{13}. The All-in-One Network\cite{14} proposes a framework with separate encoders for each weather but a generic decoder and neural architecture search across weather-specific optimized encoders. The Transweather\cite{15} using vision transformer construction has a single encoder and a decoder and learns weather-type queries to solve all adverse weather removal efficiently. Wetherdiff\cite{16} using diffusion models enables size-agnostic image restoration by using a guided denoising process. To our interest, these three studies focus on the inability of specific weather combinations to adapt to new weather types. Recently, MuSe-Net\cite{17} employs a two-branch neural network containing one multiple-environment style extraction network and one self-adaptive feature extraction network to dynamically adjust the domain shift caused by environmental changes. However, this method does not perform well in some real-world high-intensity rains with a splattering effect. In summary, multi-weather cross-view geo-localization in unseen unpredictable real weather conditions is a problem that needs to be solved urgently.

To overcome these obstacles, this paper presents MCGF, a Multi-weather Cross-view Geo-localization Framework designed to dynamically adapt to unseen weather conditions, which establishes a joint optimization between image restoration and geo-localization using denoising diffusion models. Diffusion models increasingly serve discriminative tasks such as classification and image segmentation. Inspired by its powerful modeling capability and stable training process, we utilize the diffusion model to learn the denoising process from noisy images to clean images, facilitating robust matching in the presence of multi-weather.  In image restoration, MCGF includes a shared encoder and a lightweight restoration module that prompts the backbone to provide more beneficial information to eliminate the influence of weather-specific information. In geo-localization, MCGF uses EVA-02\cite{18} as a backbone for feature extraction and uses cross-entropy loss for training and cosine distance for testing. EVA-02 is a ViT\cite{19} model obtained using a series of stable optimization methods, which allows MCGF to extract more favorable information from drone and satellite images while using fewer parameters.

Extensive experiments on University160k-WX demonstrate that MCGF achieves competitive results for geo-localization in varying weather conditions. The code will be released at https://github.com/ fengtt42/ACMMM24-Solution-MCGF.

\begin{figure*}[t]
    \centering
    \includegraphics[width=0.74\textwidth]{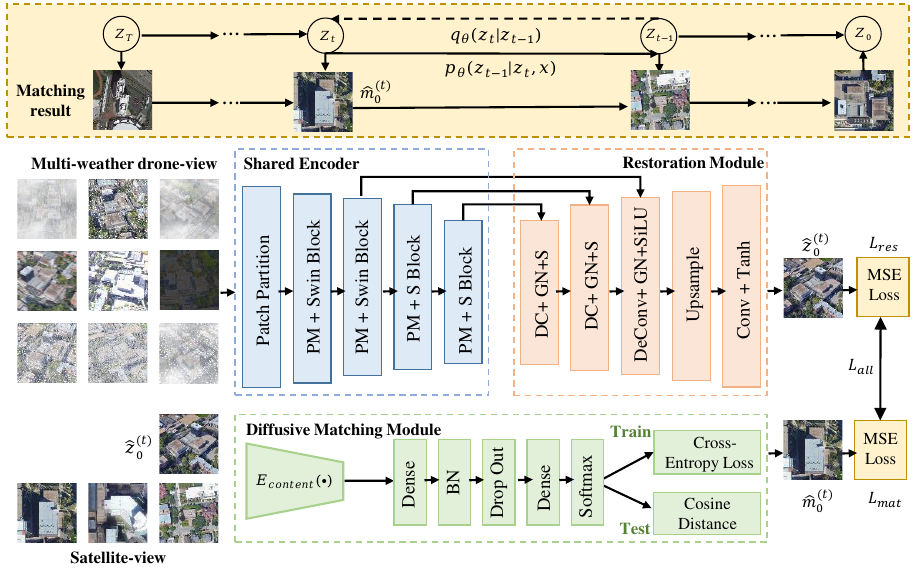}
    \caption{The overview structure of MCGF, which establishes a joint optimization between image restoration and geo-localization.}
    \label{fig2}
\end{figure*}

\section{Method}
MCGF establishes a joint optimization between image restoration and geo-localization using denoising diffusion models. In image restoration, MCGF uses a shared encoder and a lightweight restoration module to gradual denoising and obtain clearer drone-viewing images. In geo-localization, MCGF uses a diffusive matching module for cross-view matching, which can make the matching module run at multiple granularities, resulting in more accurate matching results. The overview structure of MCGF is shown in Figure \ref{fig2}.

\subsection{Denoising Diffusion Models}
The diffusion model is a probabilistic model that has attracted considerable interest in the computer vision community. It can remarkably approximate the original data distribution by gradually adding Gaussian noise to the training data and learning to reverse this diffusion process.

The forward process is a fixed Markov Chain that sequentially corrupts the data $z_0 \sim q_\theta(z_0)$ at $T$ diffusion time steps, by injecting Gaussian noise according to a variance schedule $\beta_1,...,\beta_T$. Given the clean drone-view images $z_0$, the forward process at step $t$ is defined as:
\begin{equation}
q_\theta(\mathbf{z}_t \mid \mathbf{z}_{t-1}) = \mathcal{N}(\mathbf{z}_t; \sqrt{\alpha_t} \mathbf{z}_{t-1}, \beta_t \mathbf{I})
\end{equation}
\begin{equation}
q_\theta(\mathbf{z}_{1:T} \mid \mathbf{z}_0) = \prod_{t=1}^T q_\theta(z_t \mid z_{t-1})
\end{equation}
\begin{equation}
q_\theta(\mathbf{z}_t \mid \mathbf{z}_0) = \mathcal{N}(\mathbf{z}_t; \sqrt{\bar{\alpha}_t} \mathbf{z}_0, (1 - \bar{\alpha}_t) \mathbf{I})
\end{equation}
where $\alpha_t$ and $\beta_t$ are noise schedule parameters, $\bar{\alpha}_t = \prod_{s=1}^t \alpha_s$ and $\alpha_t = 1-\beta_t$.

The reverse process attempts to remove the noise added in the forward process. The reverse process defined by the joint distribution $p_\theta(z_{0:T})$ is a Markov Chain with learned Gaussian denoising transitions starting at a standard normal prior $p_\theta(z_T) = \mathcal{N}(z_T;\mathbf{0};\mathbf{I})$. At step $t$, the reverse process is defined as:
\begin{equation}
p_\theta(\mathbf{z}_{0:T}) = p(z_T) \prod_{t=1}^T p_\theta(z_{t-1} \mid z_{t})
\end{equation}
\begin{equation}
p_\theta(\mathbf{z}_{t-1} \mid \mathbf{z}_t) = \mathcal{N}(\mathbf{z}_{t-1}; \mu_\theta(\mathbf{z}_t, t), \Sigma_\theta(\mathbf{z}_t, t))
\end{equation}

For simplicity, we assume \(\Sigma_\theta\) is a known constant, thus the reverse process simplifies to:
\begin{equation}
p_\theta(\mathbf{z}_{t-1} \mid \mathbf{z}_t) = \mathcal{N}(\mathbf{z}_{t-1}; \mu_\theta(\mathbf{z}_t, t), \sigma^2 \mathbf{I})
\end{equation}

Here the reverse process is parameterized by a neural network that estimates $\mu_\theta(\mathbf{z}_t, t)$ and $\Sigma_\theta(\mathbf{z}_t, t))$. The forward process variance schedule $\beta_t$ can be learned jointly with the model or kept constant, ensuring that $z_t$ approximately follows a standard normal distribution.

The training objective of the denoising diffusion model is to maximize the likelihood of the reverse process, which can be achieved by minimizing the variational lower bound (VLB) of the negative log-likelihood. The VLB is given by:
\begin{equation}
\mathcal{L}_{\text{VLB}} = \mathbb{E}_q \left[ -\log p_\theta(\mathbf{z}_0) + \mathcal{SD}_{KL} \right]
\end{equation}

\begin{equation}
\mathcal{SD}_{KL} = \left[\sum_{t=1}^T \mathcal{D}_{KL} \left[ q_\theta(\mathbf{z}_{t-1} \mid \mathbf{z}_t, \mathbf{z}_0) \, \| \, p_\theta(\mathbf{z}_{t-1} \mid \mathbf{z}_t) \right] \right]
\end{equation}

In practice, this can be decomposed into reconstruction error and KL divergence terms for each step, which are optimized accordingly.

\subsection{Shared Encoder}
To enhance feature representation and improve subsequent image restoration and geo-localization, we utilize the widely adopted state-of-the-art transformer-based model, Swin Transformer\cite{21}, as the shared encoder in our unified framework. The Swin Transformer is a hierarchical transformer that employs shifted windows, which restricts attention computation to non-overlapping local windows, making it adaptable for modeling at various scales. To balance computational overhead and inference speed, we select the tiny version of Swin Transformer as the default backbone.

\subsection{Restoration Module}
The restoration module utilizes a straightforward CNN-based encoder architecture, consisting of three deconvolutions, an upsampling, and a $Tanh$ activation function. It facilitates geo-localization by revealing clean features at multiple scales and produces weather-free images. We adopt a simple Mean Squared Error (MSE) as the loss function of the restoration subnetwork.
\begin{equation} 
L_{res}=\frac{1}{n} \sum_{i=1}^{n}(z_{0,i}-\hat{z}_{0,i}^t)^2 
\end{equation}
where $n$ denotes the patch size. It can minimize the pixel-wise difference between the clean image $z_{0,i}$ and the estimated weather-free image $\hat{z}_{0,i}^t$.

\subsection{Diffusive Matching Module}
{\it Feature extraction.} MCGF introduces the latest transformer-based visual representation, EVA-02, as the backbone of $E_{content}(\centerdot)$ in the network. In fact, EVA-02 has shown superior performance in most CV downstream tasks. EVA’s architecture is a vanilla ViT encoder that can be regarded as a student model, with a shape following ViT giant and the vision encoder of BEiT-3. A big dataset, consisting of several typical and openly accessible datasets with 29.6 million images in total, is used as pre-training data. After pre-training, EVA is scaled up to 1.0B parameters compared to CLIP. Based on the theory of EVA, larger CLIP-like models will provide more robust target representations for masking image modeling.

{\it Loss calculation.} Because the training set and the testing set do not overlap in terms of image-matching categories, there are many new categories in the test set. During training, since there is a large dataset, the cross-entropy loss function can better converge the model. However, for new categories in the test set, the cosine distance can solve this problem well. So the feature map extracted by $E_{content}(\centerdot)$ encoder is fed into a multilayer perceptron (MLP) to calculate the cross-entropy loss for training or cosine distance for testing. MLP includes 2 dense layers, a Batch Normalization (BN) layer, a drop out layer, and a softmax activation function. 

{\it Optimization.} MCGF contains two loss functions, one is image restoration loss $L_{res}$, and the other is matching loss $L_{mat}$. The joint optimization between image restoration and geo-localization is achieved through total loss function $L_{all}$. At every time $t$, denoised images $\hat{z}_0^t$ and matching images $\hat{m}_0^t$ are obtained, and the joint optimization is achieved by minimizing the cumulative loss $L_{all}$.

\begin{equation} 
L_{mat}=\frac{1}{n} \sum_{i=1}^{n}(m_{0,i}-\hat{m}_{0,i}^t)^2 
\end{equation}
\begin{equation} 
L_{all}=L_{res} + L_{mat}
\end{equation}
where $n$ denotes the patch size. In the gradual denoising process of the restoration module, the diffusive matching module can gradually obtain clearer drone-viewing images as input. This process enables the matching model to run at multiple granularities, resulting in more accurate matching results.

\section{Experiment}
{\it Dataset.} University160k-WX\cite{0} is a multi-weather cross-view geo-localization dataset, which extends the University-1652 dataset with extra 167,486 satellite-view gallery distractors. University160k-WX further introduces weather variants on University160k, including fog, rain, snow and multiple weather compositions. These distractor satellite-view images have a size of 1024 × 1024 and are obtained by cutting orthophoto images of real urban and surrounding areas. Multiple weathers are randomly sampled to increase the difficulty of representation learning.

{\it Implement details.} We employed the EVA-02 model, which is based on the Vision Transformer, as the backbone for diffusive matching module. This model has been trained and fine-tuned on many large vision datasets. In our experiments, we resized each input image to a fixed size of 448 × 448 pixels. During training, we used SGD as the optimizer with a momentum of 0.9 and weight decay of 5 × $10^{-4}$, with a mini-batch size of 16. The initial learning rate was set to 0.01 for the backbone layer and 0.1 for the classification layer. Our model was built using Pytorch.

{\it Evaluation metrics.} The performance of our method is evaluated by the Recall@K (R@K) and the average precision (AP). R@K denotes the proportion of correctly localized images in the top-K list, and R@1 is an important indicator. AP is equal to the area under the Precision-Recall curve. Higher scores of R@K and AP indicate better performance of the network.

\subsection{Geo-localization results} 
We train MCGF with outstanding algorithms (including LPN\cite{22}, MBEG\cite{23}, and Muse-Net\cite{17}) on the University-160k-WX train set until convergence and obtain optimal results. We test all trained models on the official unified test set provided by the competition organizer. All test results can be displayed and downloaded on the competition result submission platform. Table \ref{table1} shows that MCGF is significantly better than existing methods in all evaluation metrics. Especially compared with the latest research Muse-Net, MCGF can achieve a 67.75{\%} performance improvement in the Recall@1 indicator. MCGF shows considerable potential for geo-localization as a general framework.

\subsection{Visualization}
\begin{figure}[t]
    \centering
    \includegraphics[width=0.475\textwidth]{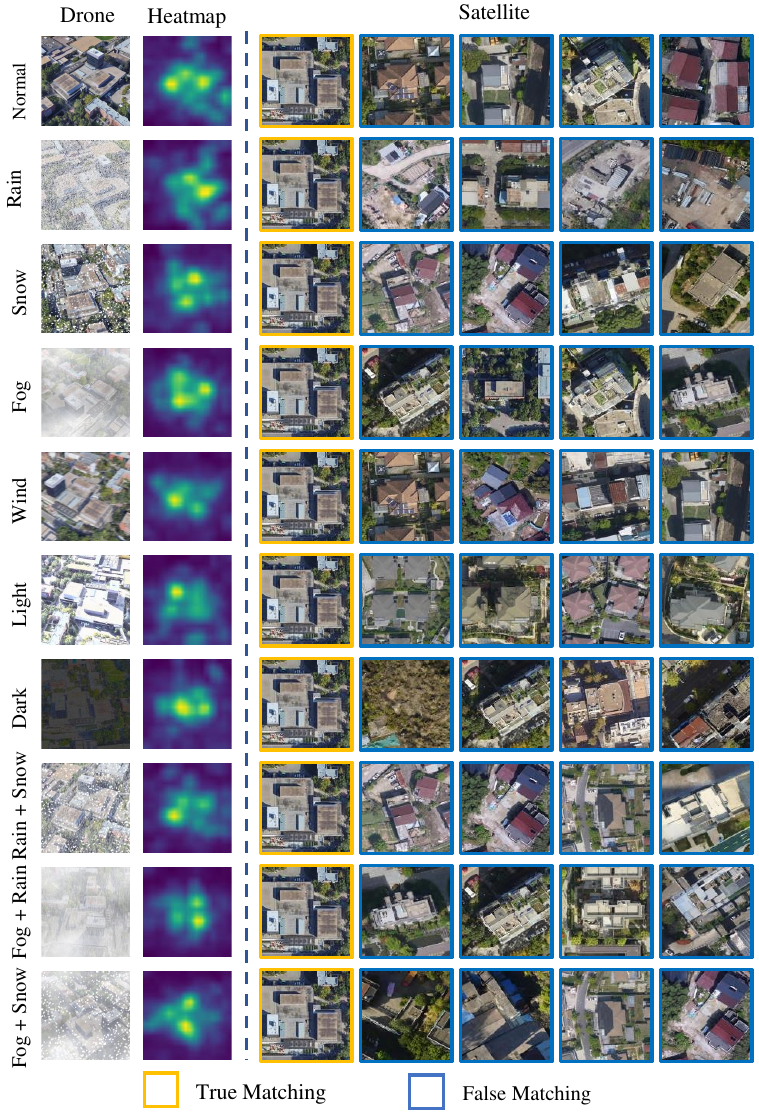}
    \caption{Visualization of heatmaps generated by our method and Top-5 matching results for a drone-view image in different conditions.}
    \label{fig3}
\end{figure}

As shown in Figure \ref{fig3}, we visualise heatmaps and Top-5 matching results generated by our method in 10 different weather conditions. Since the drone is flying around, the drone images is not only interfered by weather but also by rotational posture. Therefore, we also show the impact of drone posture changes on geo-localization in Figure \ref{fig3}. The heatmap shows that our method can accurately extract the shape and relative position of geographic targets under weather and pose interference. From the matching results shown, we observe that our model obtains the true match in the Top-1 yet the remaining matching results are inconsistent under 10 different conditions, which also indicates that the adjusted features still contain a few discrepancies.

\section{Conclusion}
This paper presents a multi-weather cross-view geo-localization framework designed to dynamically adapt to unseen weather conditions, which establishes a joint optimization between image restoration and geo-localization using denoising diffusion models. In image restoration, MCGF uses a shared encoder and a lightweight restoration module to gradual denoising and obtain clearer drone-viewing images. In geo-localization, MCGF uses a diffusive matching module for cross-view matching. The limitation of this method is that it needs to take a long time to train. For future research, it provides a joint optimization framework based on the diffusion model, which can be applied to other tasks, such as matching watermarked images, matching stained images, and matching occluded images.

\section{Acknowledgments}
This work was supported in part by the National Key Research and Development Program of China No. 2020AAA0106300, National Natural Science Foundation of China (No. 62222209, 62250008, 62102222), Beijing National Research Center for Information Science and Technology (No. BNR2023RC01003, BNR2023TD03006), China Postdoctoral Science Foundation under Grant No. 2024M751688, Postdoctoral Fellowship Program of CPSF under Grant No. GZC20240827, and Beijing Key Lab of Networked Multimedia.

\begin{table}[t]
	\caption{Matching results compared with SOTA methods.}
	\label{table1}
	\resizebox{0.38\textwidth}{!}{%
		\begin{tabular}{ccccc}
			\toprule
			Methods & R@1 & R@5 & R@10 & AP \\
			\midrule
			LPN\cite{22} & 7.98 & 10.25 & 11.21 & 8.49 \\
			MBEG\cite{23} & 26.17 & 32.84 & 35.32 & 29.32\\
			Muse-Net\cite{17} & 50.48 & 63.19 & 67.34 & 53.27 \\
			MCGF(ours) & 84.68 & 91.36 & 93.00 & 88.71\\
			\bottomrule
		\end{tabular}%
	}
\end{table}

\enlargethispage{-110mm}
\bibliographystyle{ACM-Reference-Format}
\bibliography{MCGF}

\end{document}